%% file: acl_latex.tex
\pdfoutput=1

\documentclass[11pt]{article}
\usepackage[usenames,dvipsnames]{xcolor}
\usepackage{mathtools}
\usepackage{stmaryrd}
\usepackage{amssymb}
\usepackage{bm}
\usepackage{bbm}
\usepackage{tabularx} 

\usepackage{acl2023}
\usepackage{multirow}

\usepackage{times}
\usepackage{latexsym}
\usepackage{graphicx}
\usepackage[thinlines]{easytable}
\usepackage{booktabs}
\usepackage{amssymb}
\usepackage{amsmath}
\usepackage{caption}
\usepackage{subcaption}
\usepackage{listings}
\usepackage{multicol}

\usepackage[T1]{fontenc}

\usepackage[utf8]{inputenc}

\usepackage{microtype}

\usepackage{xparse}
\usepackage{listings}
\usepackage{algorithm}
\usepackage{algpseudocode}
\usepackage[most]{tcolorbox}

\newcommand{\ours }{\textsc{IncSchema}}
\newcommand{\dataset}{\textsc{ODiN}}

\usepackage{colortbl}
\definecolor{LightGray}{gray}{0.9}
\definecolor{aliceblue}{rgb}{0.94, 0.97, 1.0}
\definecolor{almond}{rgb}{1.0, 0.92, 0.8}
\newcolumntype{a}{>{\columncolor{aliceblue}}c}

\lstdefinestyle{mystyle}{
    backgroundcolor=\color{aliceblue},   
    basicstyle=\ttfamily\footnotesize,
    breakatwhitespace=false,         
    breaklines=true,                 
    captionpos=b,                    
    keepspaces=true,                 
    numbers=none,                    
    showspaces=false,                
    showstringspaces=false,
    showtabs=false,                  
    tabsize=2
}

\title{
Open-Domain Hierarchical Event Schema Induction by Incremental Prompting and Verification}

\author{Sha Li$^1$,  Ruining Zhao$^1$, Manling Li$^1$, Heng Ji$^1$, Chris Callison-Burch$^2$,  Jiawei Han$^1$ \\
$^1$University of Illinois at Urbana-Champaign \\
$^2$ University of Pennsylvania\\
\texttt{\{shal2, ruining9, manling2, hengji, hanj\}@illinois.edu} \\
\texttt{ccb@seas.upenn.edu}
}

\begin{document}
\maketitle
\input{0_abstract}

\input{1_introduction}

\input{2_problem}

\input{3_model}

\input{4_exp}

\input{5_related}

\input{6_conclusion}

\input{7_limitations}

\section*{Acknowledgement}
We thank the anonymous reviewers for their helpful suggestions. 
This research is based upon work supported by U.S. DARPA KAIROS Program No. FA8750-19-2-1004, the DARPA LwLL Program (contract FA8750-19-2-0201), the IARPA HIATUS Program (contract 2022-22072200005), and the NSF (Award 1928631). Approved for Public Release, Distribution Unlimited.  The views and conclusions contained herein are those of the authors and should not be interpreted as necessarily representing the official policies, either expressed or implied, of DARPA, or the U.S. Government. The U.S. Government is authorized to reproduce and distribute reprints for governmental purposes notwithstanding any copyright annotation therein.

\bibliography{anthology, custom}
\bibliographystyle{acl_natbib}

\appendix

\input{appendix}

\end{document}

%% file: 0_abstract.tex
\begin{abstract}
   Event schemas are a form of world knowledge about the typical progression of events. 
   Recent methods for event schema induction use information extraction systems to construct a large number of event graph instances from documents, and then learn to generalize the schema from such instances. 
   In contrast, we propose to treat event schemas as a form of commonsense knowledge that can be derived from large language models (LLMs). 
   This new paradigm greatly simplifies the schema induction process and 
   allows us to handle both hierarchical relations and temporal relations between events in a straightforward way. 
   Since event schemas have complex graph structures, we design an \textit{incremental prompting and verification} method \ours\ 
   to break down the construction of a complex event graph into three stages: event skeleton construction, event expansion, and event-event relation verification.
   Compared to directly using LLMs to generate a linearized graph, \ours\ can generate large and complex schemas with 7.2\% F1 improvement in temporal relations and 31.0\% F1 improvement in hierarchical relations. In addition, compared to the previous state-of-the-art closed-domain schema induction model, human assessors were able to cover $\sim$10\% more events when translating the schemas into coherent stories and rated our schemas 1.3 points higher (on a 5-point scale) in terms of readability. 
  \footnote{Code and \dataset\ dataset available at \url{https://github.com/raspberryice/inc-schema}.}
\end{abstract}

%% file: 1_introduction.tex
\section{Introduction}
\begin{figure}
    \centering
    \includegraphics[width=\linewidth]{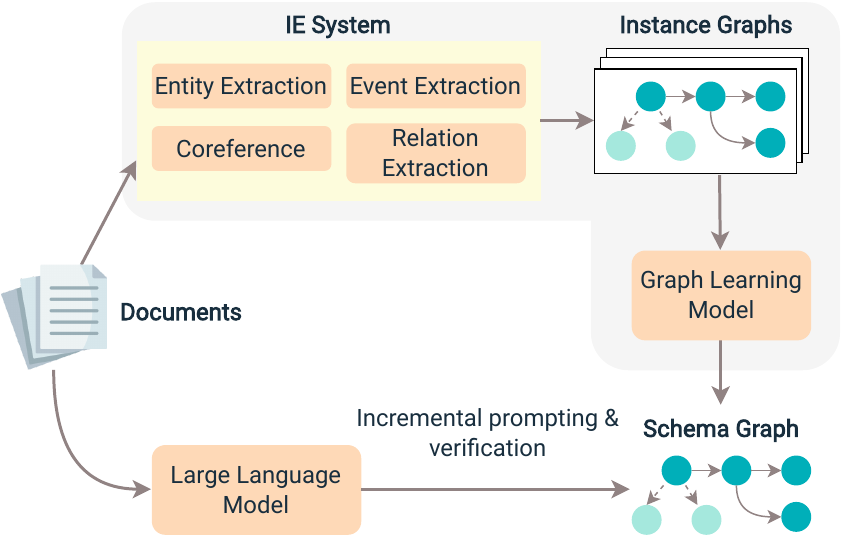}
    \caption{A comparison between the instance-based schema induction pipeline (gray background) and our \ours\ approach. By directly prompting LLMs to construct the schema graph, our framework is conceptually simpler, open-domain, extensible, and more interpretable.}
    \label{fig:intro}
\end{figure}
Schemas, defined by \cite{schank1975scripts} as ``a predetermined, stereotyped sequence of actions that defines a well-known situation'', are a manifestation of world knowledge. With the help of schemas, a model can then infer missing events such as a person must have ``been within contact with a pathogen'' before the event ``the person was sent to the hospital for treatment'' and also predict that if a large-scale incident happened, this might trigger an ``investigation of the source of the pathogen''.

To automate schema creation, two mainstream approaches are to  
learn from manually created reference schemas
or learn from large amounts of event instances automatically extracted from documents.
Manual creation of complex hierarchical schemas requires expert annotation, which is not scalable\footnote{Online crowdsourced resources such as WikiHow only contain simple linear schemas.}.
On the other hand, instance-based schema induction methods~\cite{li-etal-2020-connecting,li-etal-2021-future,jin-etal-2022-event,dror2022zeroshotschema} rely on complicated preprocessing\footnote{To create an event instance graph, typical steps include entity extraction, entity-entity relation extraction, event extraction, coreference resolution, and event-event relation extraction.} to transform documents into instance graphs for learning. 
Moreover, supervised information extraction systems~\cite{ji2008refining,Lin2021} are domain-specific and suffer from error propagation through multiple components, which makes the downstream schema induction model closed-domain and low in quality. 

\begin{figure*}[th]
    \centering
    \includegraphics[width=\linewidth]{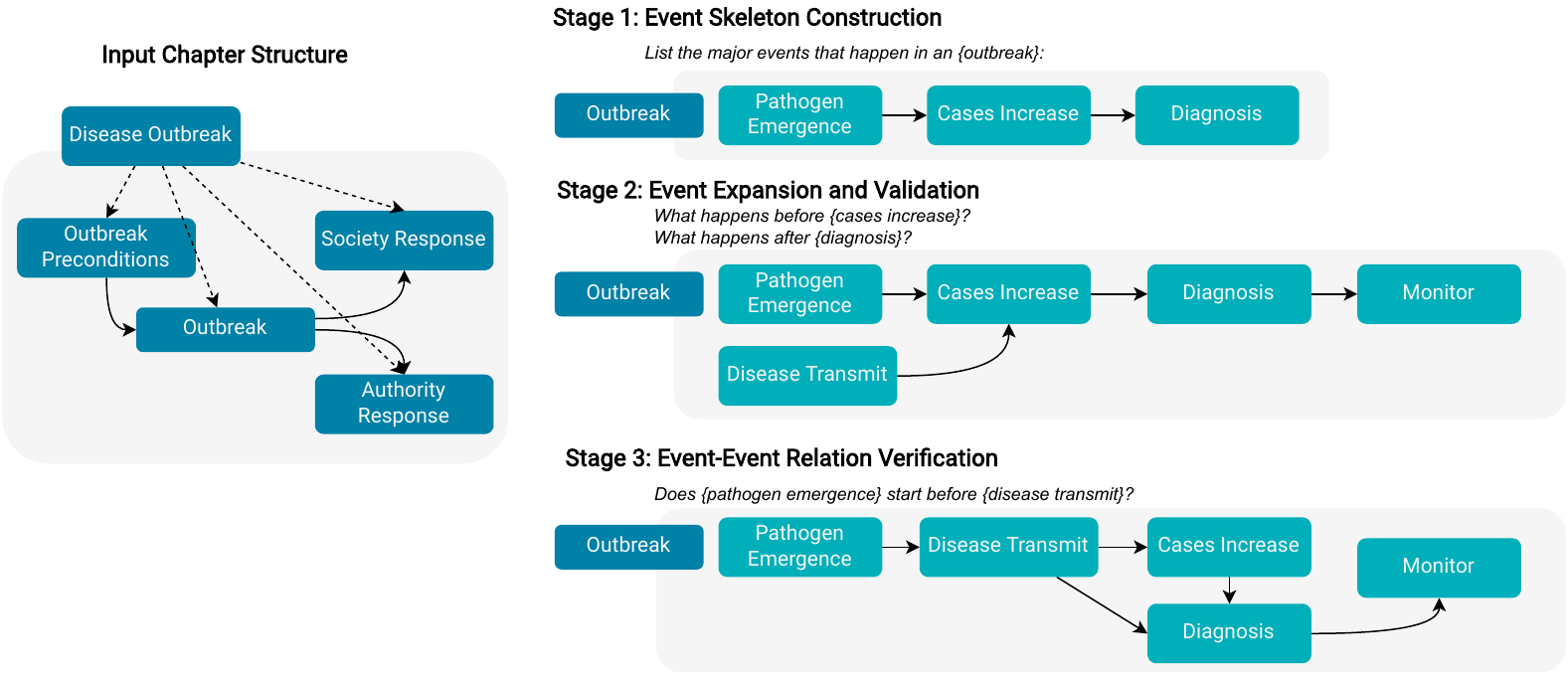}
    \caption{To create the schema for a given scenario, our model follows 3 rounds of operation: (1) \textit{event skeleton construction} where we ask the LLM to list the important events; (2) \textit{event expansion} to discover more related events for each existing event; \textit{event-event relation verification} where we update the event-event relations based on the LLM's answers to questions about each event pair.}
    \label{fig:overview}
\end{figure*}

Tracing back to the original definition of schemas, we observe that ``stereotyped sequences of events'' or ``the typical progression of events'' can be viewed as \textit{commonsense knowledge} that can be implicitly learned by training on large corpora. Through the language modeling objective, models can pick up which events statistically frequently co-occur and how their relationship is typically described. 
More recently, large language models (LLMs) such as GPT3~\cite{Brown2020GPT3} and PaLM~\cite{Chowdhery2022PaLM} have shown impressive zero-shot performance on closely-related commonsense reasoning tasks such as goal-step reasoning~\cite{zhang-etal-2020-reasoning} and temporal reasoning\footnote{\url{https://github.com/google/BIG-bench/tree/main/bigbench/benchmark_tasks/temporal_sequences}}.

By utilizing LLMs to directly prompt for schematic knowledge, our approach is \textit{open-domain}, \textit{extensible} and \textit{more interpretable} for humans.
Given a new scenario name, our model only requires lightweight human guidance in providing some top-level chapter structure (as shown in the left of Figure \ref{fig:overview}) and can produce the entire schema in under an hour whereas instance-based methods require months to collect the data and retrain the IE system for new domains. Our model is \textit{extensible} and can support new types of event-event relations by adding new prompt templates. To showcase this, in addition to the temporal relation between events which is the focus of prior work, we also account for the different event granularities by supporting hierarchical relations between events (for example, a \texttt{physical conflict} could happen as a part of a \texttt{protest}). 
Finally, by representing events with free-form text instead of types and organizing them into a hierarchy, our generated schemas are considered more interpretable.

We find that directly asking LLMs to generate linearized strings of schemas leads to suboptimal results due to the size and complexity of the graph structure.
To solve this problem, we design an \textit{incremental prompting and verification} scheme to break down the construction of a complex event graph schema into three major stages: event skeleton construction, event expansion, and event-event relation verification.
As shown in Figure~\ref{fig:overview}, each stage utilizes templated prompts (\texttt{What happens before \underline{cases increase}?}) which can be instantiated either with the scenario name or the name of a previously generated event.

The key contributions of this paper are:
\begin{itemize}
    \item We propose a framework \ours\ for inducing complex event schemas by treating the task as knowledge probing from LLMs. Compared to previous approaches that rely on the creation of event instance graphs, our method greatly simplifies the process and as a result, is not confined to the working domain of any IE system. 
    \item We extend the expressive power of event schemas by inducing hierarchical relations and temporal relations between events at the same time. 
    Our modularized prompting framework allows us to support a new type of event-event relation easily, whereas prior work~\cite{zhou-etal-2022-show,dror2022zeroshotschema} required specialized pipelines or components.

    \item We verify the effectiveness of our framework on two complex schema datasets: \dataset, an Open-Domain Newswire schema library, and RESIN-11~\cite{du-etal-2022-resin}. Compared to directly generating the schema using a linearized graph description language~\cite{sakaguchi-etal-2021-proscript-partially}, \ours\ shows 7.2\% improvement in temporal relation F1 and 31.0\% improvement in hierarchical relation F1. 
\end{itemize}

%% file: 2_problem.tex
\section{Task Overview}
Given a scenario name, a schema depicts the \textit{general progression} of events within that scenario.

Following \cite{li-etal-2021-future}, we consider the schema to be a graph structure of events. 
We output a schema graph of event nodes and event-event relation edges, including temporal relations and hierarchical relations.

Since our algorithm is designed to be open-domain, we represent each event $e$ with a description string such as ``\texttt{A person shows early symptoms of the disease}'' instead of a type from a restricted ontology 
 (e.g., \texttt{Illness}).  Description strings are more flexible in representing different granularities of events and are more informative.
 It is noteworthy that event descriptions in a schema should be general, instead of a specific instance, such as ``\texttt{John had a mild fever due to COVID}''.

In addition, we support the representation of chapters, which are ``a collection of events that share the same theme and are connected in space-time''.
When a high-level chapter structure $G_c$ (as shown in the left side of Figure \ref{fig:overview}) is available, we condition on the given chapters to guide the schema generation process. Chapters are also treated as events and can potentially have temporal relations between them. Every other event must be a descendant of a chapter event. If no chapter structure is available, we create a single chapter from the scenario name.

%% file: 3_model.tex
\section{Our Approach}
Leveraging LLMs to directly generate the full schema graph is challenging due to the size and complexity of schemas. Thus, we divide our schema induction algorithm \ours\ into three stages as depicted in Figure \ref{fig:overview}.
Starting from the scenario node or one of the chapter nodes, the skeleton construction stage first produces a list of major events that are subevents of the scenario (chapter) following sequential order.
For each generated event, we expand the schema graph to include its temporally-related neighbors and potential children in the event expansion stage. For each pair of events, we further rescore their temporal and hierarchical relation probability in the relation verification stage to enrich the relations between events.

\subsection{Retrieval-Augmented Prompting}
\label{sec:retrieval}
To make the model more informed of how events are typically depicted in news, we introduce a retrieval component %
to guide LLMs to focus on scenario-related passages. The key difficulty of schema induction is to generalize from multiple passages and reflect the ``stereotyped sequence of events'' instead of providing concrete and specific answers. We, therefore,  retrieve multiple passages each time and ask the model to provide a generalized answer that is suitable for all passages. 

To build a document collection containing typical events of the given scenario, we leverage its Wikipedia category page and retrieve the reference news articles of each Wikipedia article under the category, as detailed in Appendix~\ref{sec:appendix_retrieval}. 
With such a document collection, for each prompt, %
we are able to use the description of the event as the query and retrieve $k=3$ passages based on state-of-the-art document retrieval system 
TCT-ColBERT~\cite{lin-etal-2021-batch}.
The input to the LM is structured as follows: 

\begin{tcolorbox}[colback=blue!5!white,colframe=Emerald!80!black,title=Retrieval-Augmented Prompt]
  Based on the following passages \{\texttt{retrieved passages}\}, \\
  \{\texttt{prompt}\}
\end{tcolorbox}
Providing more than one passage is critical as we want the model to produce a \textit{generalized} response instead of a specific response that only pertains to one event instance.

\subsection{Event Skeleton Construction}
\label{sec:skeleton}

We use the following prompt to query the LM about events that belong to the chapter $c$: 
\begin{tcolorbox}[colback=blue!5!white,colframe=Emerald!80!black,title=Event Skeleton Prompt]
 \{evt.name\} is defined as "\{evt.description\}". 
List the major events that happen in the \{evt.name\} of a \{scenario\}:
\end{tcolorbox}
This typically gives us a list of sentences, which is further translated into a linear chain of event nodes by treating each sentence as an event description and regarding the events as listed in temporal order. 
To assign a name to each event for easier human understanding, 
 we leverage the LLM again with in-context learning using 10 \{description, name\} pairs such as \{\texttt{Disinfect the area to prevent infection of the disease, Sanitize}\} (the complete list of in-context examples is in Appendix \ref{sec:appendix_prompt}). 

\subsection{Event Expansion and Validation}
\label{sec:event_validation}
Given an event $e$ (such as \textit{Cases Increase} in Figure \ref{fig:overview}), we expand the schema by probing for its connected events in terms of temporal and hierarchical relations using prompts as below:
\begin{tcolorbox}[colback=blue!5!white,colframe=Emerald!80!black,title=Event Expansion Prompt]
What happened during "\{evt.description\}"? List the answers: 
\end{tcolorbox}
(See Appendix \ref{sec:appendix_prompt} for a full list of prompts used.) 
Every sentence in the generated response will be treated as a candidate event.

For every candidate event $e'$ (such as \textit{DiseaseTransmit} in Figure \ref{fig:overview}), we perform a few validation tests as listed below. The event is only added to the schema when all the tests pass. 

   \paragraph{Duplication Test} To check if a new event is a duplicate of an existing event, we use both embedding similarity computed through cosine similarity of SBERT embeddings~\cite{reimers-gurevych-2019-sentence}\footnote{We use the \texttt{all-MiniLM-L6-v2} model.} and string similarity using Jaro-Winkler similarity~\cite{winkler1990string}. If the event description, event name, or the embedding of the event description is sufficiently similar to an existing event in the schema, we will discard the new event. 
    \footnote{This threshold is determined empirically, and we set it to 0.9 for Jaro-Winkler string similarity, 0.85 for embedding cosine similarity.} 
    \paragraph{Specificity Test} When we augment the prompt with retrieved documents, at times the model will answer the prompt with details that are too specific to a certain news article, for instance, include the time and location of the event. The specificity test seeks to remove such events. 
    We implement this by asking the LLM ``Does the text contain any specific names, numbers, locations, or dates?'' and requesting a yes-no answer. We use 10 in-context examples to help the LLM adhere to the correct answer format and understand the instructions. 
    
    \paragraph{Chapter Test}
    For the chapter assignment test, we present the name and the definition of the chapter event $c$ and the target event $e'$ respectively, then ask ``Is $e'$ a part of $c$? ''. If the answer is ``yes'', we keep the event $e'$.

If a new event $e'$ passes validation, we assign a name to the event following the same procedure as in Section \ref{sec:skeleton}.

\begin{table*}[t]
    \centering
    \small 
    \setlength\tabcolsep{2pt}
    \setlength\extrarowheight{0pt}
    \begin{tabular}{l|m{16em} | c | c | c}
    \toprule
    Relation & Allen's base relations &  $e_1$ starts before $e_2$? & $e_1$ ends before $e_2$? & $e_1$ duration longer than $e_2$?  \\
    \midrule 
       $e_1 \prec e_2 $  &  $e_1$ precedes $e_2$, $e_1$ meets $e_2$ & Yes & Yes & - \\
       $e_1 \succ e_2 $  &  $e_1$ is preceded by $e_2$, $e_1$ is met by $e_2$ & No & No & - \\
       $e_1 \subset e_2$  &  $e_1$ starts $e_2$, $e_1$ during $e_2$, $e_1$ finishes $e_2$ & No & Yes & No \\
        $e_1 \supset e_2$  &  $e_1$ is started by $e_2$, $e_1$ contains $e_2$,  $e_1$ is finished by $e_2$ & Yes & No & Yes \\
         $e_1 \parallel e_2 $ & $e_1$ overlaps with $e_2$, $e_1$ is equal to $e_2$ & Yes & No & No  \\
          $e_1 \parallel e_2 $ & $e_1$ is overlapped by $e_2$ & No & Yes & Yes   \\
       \bottomrule 
    \end{tabular}
    \caption{Schema event-event relations, the correspondence to Allen's interval algebra, and the related relation questions. In the case of temporal overlap (last two rows), we refrain from adding an edge between the two events.}
    \label{tab:relations}
\end{table*}
\subsection{Event-Event Relation Verification}
Although the prompts from the previous step naturally provide us with some relations between events (the answer to ``What are the steps in $e$?'' should be subevents of $e$), such relations may be incomplete or noisy.
To remedy this problem, for every pair of events $(e_1, e_2)$ in the same chapter, we verify their potential temporal/hierarchical relation.

A straightforward way to perform verification would be to ask questions such as ``Is $e_1$ a part of $e_2$?'' and ``Does $e_1$ happen before $e_2$?''.
Our pilot experiments show that this form of verification leads to sub-optimal results in two aspects: (1) \textit{relation confusion}: the language model will predict both $e_2 \prec e_1$ and $e_1 \subset e_2$; and
(2) \textit{order sensitivity}: the language model tends to return ``yes'' for both ``Does $e_1$ happen before $e_2$?'' and ``Does $e_1$ happen after $e_2$?''.

To solve this \textit{relation confusion} problem, inspired by Allen interval algebra~\cite{Allen1983MaintainingKA} and the neural-symbolic system in \cite{zhou-etal-2021-temporal}, we decompose the decision of a temporal relation into questions about start time, end time, and duration.
In addition, following HiEve~\cite{glavas-etal-2014-hieve}, we define the hierarchical relation as spatial-temporal containment. Thus a necessary condition for a hierarchical relation to hold between $e_1$ and $e_2$ is that the time period of $e_1$ contains $e_2$. This allows us to make decisions about temporal relations and hierarchical relations jointly using the three questions as shown in Table \ref{tab:relations}. 

\begin{tcolorbox}[colback=blue!5!white,colframe=Emerald!80!black,title=Relation Verification Prompt]
Does ``\{e1.description\}" start before ``\{e2.description\}"? \\
Answer yes, no, or unknown.
\end{tcolorbox}

For each question, to obtain the probability of the answers, we take the log probability of the top 5 tokens\footnote{At the time of writing, OpenAI API only supports returning the log probability of a maximum of 5 tokens.} in the vocabulary and check for the probability predicted for ``yes'', ``no'' and ``unknown'' tokens. 

To handle the order sensitivity, we average the scores obtained from the different orderings (``Does $e_1$ start before $e_2$?'' and ``Does $e_2$ start before $e_1$'') and different prompts (``Does $e_1$ start before $e_2$?'' and ``Does $e_2$ start after $e_1$?'').

After obtaining the response for start time, end time, and duration questions, we only keep edges that have scores higher than a certain threshold for all of the three questions. 

Since our temporal edges were only scored based on the descriptions of the event pair, we need to remove loops consisting of more than 2 events, ideally with minimal changes, to maintain global consistency. This problem is equivalent to the problem of finding the \textit{minimal feedback arc set}, which is shown to be NP-hard. We adopt the greedy algorithm proposed in ~\cite{eades1993fast} using the previously predicted probabilities as edge weights to obtain a node ordering. Based on this ordering we can keep all edges directionally consistent. %
The detailed algorithm is provided in Appendix \ref{sec:appendix_a}.
Finally, to simplify the schema, we perform transitive reduction on the relation and hierarchy edges respectively.

%% file: 4_exp.tex
\section{Experiments}
\label{sec:exp}
We design our experiments based on the following three research questions: 
\paragraph{Q1: Hierarchical Schema Quality} Can our model produce high-quality event graph schemas with both temporal and hierarchical relations? 
\paragraph{Q2: Interpretability} Is our model's output more interpretable than prior instance-based schema induction methods?
\paragraph{Q3: Model Generalization} Can our model also be applied to everyday scenarios as in \cite{sakaguchi-etal-2021-proscript-partially}?
\subsection{Dataset}
RESIN-11~\cite{du-etal-2022-resin} is a schema library targeted at 11 newsworthy scenarios and includes both temporal and hierarchical relations between events. However, RESIN-11 is still quite heavily focused on attack and disaster-related scenarios, so we expand the coverage and create a new Open-Domain Newswire schema library \dataset\, which consists of 18 new scenarios, including coup, investment, and health care. 
The complete list of scenarios is in Appendix \ref{sec:appendix_scenarios}. 

Upon selecting the scenarios, we collected related documents from Wikipedia (following the procedure described in Section \ref{sec:retrieval}) and create the ground truth reference schemas by asking human annotators to curate the schemas generated by our algorithm by referring to the news reports of event instances. Human annotators used a schema visualization tool
\footnote{\url{https://schemacuration.colorado.edu/}} 
to help visualize the graph structure while performing curation. 
Curators were encouraged to (1) add or remove events; (2) change the event names and descriptions; (3) change the temporal ordering between events; and (4) change the hierarchical relation between events. After the curation, the schemas were examined by linguistic experts. 
We present the statistics of \dataset\ along with RESIN-11 and ProScript in Table \ref{tab:dataset}.

\subsection{Evaluation Metrics}

For automatic evaluation of the schema quality against human-created schemas, we adopt \textbf{Event F1} and \textbf{Relation F1} metrics.
Event F1 is similar to the Event Match metric proposed in \cite{li-etal-2021-future} but since here we are generating event descriptions instead of performing classification over a fixed set of event types, we first compute the similarity score $s$ between each generated event description and ground truth event description using cosine similarity of SBERT embeddings~\cite{reimers-gurevych-2019-sentence}. Then we find the maximum weight matching assignment $\phi$ between the predicted events $\hat E$ and the ground truth events $E$ by treating it as an assignment problem between two bipartite graphs\footnote{\url{https://en.wikipedia.org/wiki/Assignment_problem}, we use the implementation from \texttt{scipy}(\url{https://docs.scipy.org/doc/scipy/reference/generated/scipy.optimize.linear_sum_assignment.html})}.

Based on the event mapping $\phi$, we further define \textbf{Relation F1} metrics for temporal relations and hierarchical relations respectively. Note that this metric only applies to events that have a mapping.

\subsection{Implementation Details}
For both our model and the baseline, we use the GPT3 model \texttt{text-davinci-003} through the OpenAI API. 
We set the temperature to 0.7 and \texttt{top\_p} to 0.95. 

For \ours\, we set the minimum number of events within a chapter to be 3 and the maximal number of events to be 10. During the event skeleton construction stage, if the response contains less than 3 sentences, we will re-sample the response. 
Once the number of events within a chapter reaches the maximal limit, we will not add any more new events through the event expansion stage.

We set the threshold for the duplication test to be 0.9 for Jaro-Winkler string similar OR 0.85 for cosine similarity between SBERT embeddings. 
For the shorter event name, we also check if the Levenshtein edit distance is less than 3. 
For the event-event relation verification, we set the threshold for the start time and end time questions to be 0.2 and the threshold for the duration question to be 0.7.

\begin{table}[t]
    \centering
    \small 
    \begin{tabular}{l|c c c c}
    \toprule 
       Dataset  &  \# Scenarios & \# Event & \#  Temp. & \# Hier. \\
    \midrule 
     RESIN-11 &  11 & 579 & 381 & 603 \\ 
     \dataset & 18  & 593 & 398 & 569\\
     ProScript &  2077 & 14997 & 13946 & 0\\
     \bottomrule 
    \end{tabular}
    \caption{Dataset statistics. RESIN-11 and \dataset\ both focus on hierarchical schemas for newsworthy scenarios. ProScript is a collection of small-sized schemas for everyday scenarios.
    }
    \label{tab:dataset}
\end{table}

\begin{table*}[t]
    \centering
    \small 
    \begin{tabular}{l |c c a c c a c c a }
    \toprule 
         &  \multicolumn{3}{c}{Event} & \multicolumn{3}{ c}{Temp. Relation} & \multicolumn{3}{c}{ Hier. Relation}  \\ 
         Model  & Prec & Recall & F1 & Prec & Recall & F1& Prec & Recall & F1 \\
         \midrule 
         GPT-DOT &  80.0 & 30.7 & \textbf{41.8} & 24.7 & 8.31 & 11.2  
         & 11.1 & 13.7 & 12.0\\
         \ours &  39.7 & 49.3 & 41.7 & 13.8 & 
         14.8 & \textbf{13.5} & 39.3 & 38.7 & \textbf{38.9} \\ 
         \bottomrule 
    \end{tabular}
    \caption{Schema induction evaluation on RESIN-11 scenarios. Results are shown in \%.}
    \label{tab:exp-resin11}
\end{table*}

\begin{table*}[t]
    \centering
    \small 
    \begin{tabular}{l |c c a c c a c c a }
    \toprule 
         &  \multicolumn{3}{c}{Event} & \multicolumn{3}{c}{Temp. Relation} & \multicolumn{3}{c}{ Hier. Relation} \\ 
         Model  & Prec & Recall & F1 & Prec & Recall & F1 & Prec & Recall & F1  \\
         \midrule 
         GPT-DOT & 85.2 & 35.2 & 47.4 & 34.0 & 18.0 & 20.9 & 15.4 & 19.3 & 17.5\\
         GPT-DOT-Instruct & 85.4 & 34.4 & 46.7 & 33.8 & 17.8 & 20.8 & 16.7 & 18.4 & 16.3\\
         GPT-DOT-StepByStep & 95.3 & 27.1 & 41.2 & 49.5 & 21.8 & 25.9 & 13.0 & 18.1 & 14.9 \\
         \ours & 45.1 & 72.1 & \textbf{53.3} & 27.5 & 29.6 & 28.1 & 49.3 & 48.0 & 48.5  \\ 
         - No retrieval & 41.6 & 73.2 & 50.7 & 28.5 & 29.3 & \textbf{28.3} & 48.2 & 45.6 & 46.7 \\
         - No decompose & 44.8 & 71.1 & 52.4 & 25.6 & 22.7 & 23.5 & 49.6 & 50.2 & \textbf{49.7} \\
         \bottomrule 
    \end{tabular}
    \caption{Schema induction evaluation on \dataset\ scenarios. Results are shown in \%.}
    \label{tab:exp-phase2b}
\end{table*}

\begin{figure*}[th]
    \centering
    \includegraphics[width=\linewidth]{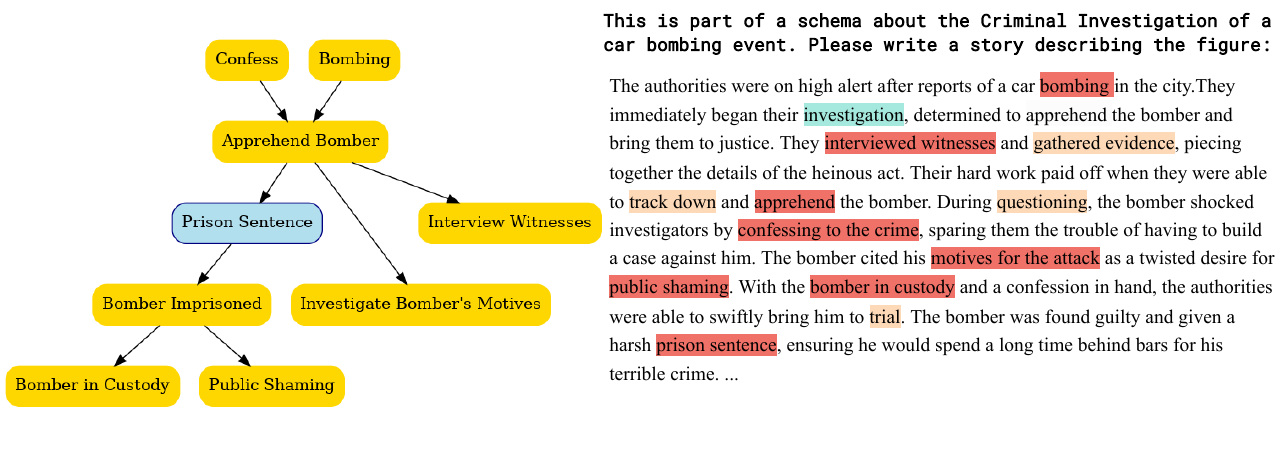}
    \caption{One example response from the schema interpretability human assessment. On the left we show the subevents of the \texttt{Criminal Investigation} chapter produced by our model. On the right is the human-written story describing the schema. We highlight the events that match the chapter in blue, events that appear in the schema in red and additional events in pink.}
    \label{fig:human_assessment_example}
\end{figure*}
\subsection{Q1: Hierarchical Schema Quality}
\label{sec:complex-schema}
We test our algorithm's ability to induce complex hierarchical schemas for news scenarios in RESIN-11~\cite{du-etal-2022-resin} and our new Open-Domain Newswire schema library \dataset. 

We compare our model against a different prompt formulation method using the DOT graph description language as purposed by~\cite{sakaguchi-etal-2021-proscript-partially}  (\textbf{GPT-DOT}). This method requires the LLM to generate all events and event-event relations in a single pass. To inform the model of the DOT language format, we use one in-context example converted from the Chemical Spill ground truth schema (the prompt is shown in Appendix \ref{sec:appendix_prompt}). During inference, we will input the scenario name and the chapter structure.

We show our results on the RESIN-11 dataset in Table \ref{tab:exp-resin11} and the results for \dataset\ in Table \ref{tab:exp-phase2b} \footnote{GPT results in Table are shown by averaging score of 5 runs.}.

Compared to our incremental prompting procedure, GPT-DOT generally outputs fewer events (10.11 events for GPT-DOT VS 52.6 events for \ours\ on \dataset), which leads to high precision but low recall. 
While the generated events from GPT-DOT are still reasonable, the real deficiency of this formulation is its inability to identify hierarchical relations, especially when hierarchical relations co-exist with temporal relations. 

To test if using an in-context learning prompt is the reason for low performance, we also experiment with an instruction-style prompt (GPT-DOT-Instruct) that explains the task and output format in detail and a step-by-step reasoning prompt (GPT-DOT-StepByStep) that allows the model to output parts of the schema separately (and we will merge them together). For the \dataset\ dataset, we find that the different prompt styles do not vary much except for improved temporal relation F1 when we use the step-by-step formulation.

Compared with the variants of our model, we can see that the retrieval component helps improve event generation quality and the question decomposition strategy can greatly improve temporal relation F1.

Since RESIN-11 schemas were created without referencing any automatic results, the scores on RESIN-11 are generally lower than that of \dataset. However, on both datasets, our method can generally outperform GPT-DOT.

\subsection{Q2: Schema Interpretability}
\label{sec:interpretability}
To be able to compare our schemas side-by-side with previous work that assumed a limited ontology, we conduct a human evaluation that focuses on the interpretability of the induced schemas.

Human assessors are presented with the scenario name and a subgraph from the schema induction algorithm's output. We then ask the assessor to write a coherent short story by looking at the graph and indicate which events were included in their story. An example of the subschema and the story is shown in Figure \ref{fig:human_assessment_example}.
After they complete the story writing task, they will be asked to rate their experience from several aspects on a 5-point Likert scale. 
The human assessment interface is shown in Appendix \ref{sec:appendix_interface}.

We compare against the state-of-the-art closed-domain schema induction method \textbf{DoubleGAE}~\cite{jin-etal-2022-event}. DoubleGAE is an example of the instance-based methods that rely on IE: the nodes in the schema graph are typed instead of described with text. 

In Table \ref{tab:storytelling_task} we show the results for the story writing task. We observe that human assessors can compose a longer story with higher event coverage when presented with our schemas while taking roughly the same time. 

\begin{figure}
    \centering
    \includegraphics[width=.8\linewidth]{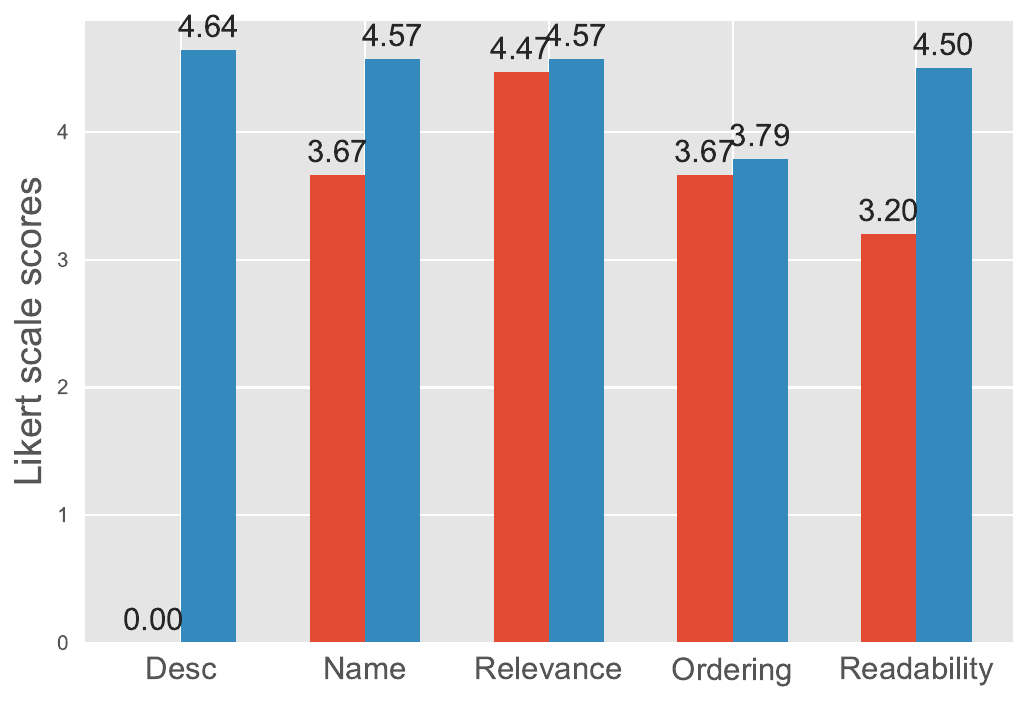}
    \caption{Human assessment of schema quality and interpretability from different aspects. Results for Double-GAE are shown in {\color{red}red} and our approach \ours\ in {\color{blue} blue}. Double-GAE does not produce event descriptions thus we omit the description helpfulness question.}
    \label{fig:human_eval}
\end{figure}

In the post-task questionnaire, as shown in Figure \ref{fig:human_eval}, the human assessors on average strongly agreed that the event names and event descriptions produced by our model were helpful and thought that our schemas were easier to understand compared to the baseline (4.50 vs 3.20 points). Both schemas contained events that were highly relevant to the scenario and the temporal ordering in the schemas was mostly correct. 
\begin{table}[t]
    \centering
    \small 
    \begin{tabular}{c|c c c}
    \toprule 
    Model & Coverage$\uparrow$ & Len(words)$\uparrow$ & Time(mins)$\downarrow$ \\
    \midrule 
    Double-GAE &  79.8 & 9.62 & 0.998 \\
     \ours    & 89.7 & 15.53 &   1.137 \\
     \bottomrule 
    \end{tabular}
    \caption{Human performance on the storytelling task using schemas from DoubleGAE and our algorithm \ours. The length and time measurements are averaged over the number of events in the schemas. Coverage is shown in percentage.
     }
    \label{tab:storytelling_task}
\end{table}

\subsection{Q3: Model Generalization}
For this experiment, we use Proscript~\cite{sakaguchi-etal-2021-proscript-partially} as our dataset.
Schemas in Proscript are typically short (5.45 events on average) and describe everyday scenarios. 
Proscript schemas only contain temporal relations and have no chapter structure, so we include the first two events (by topological sorting) as part of the prompt. We show the results in Table \ref{tab:exp-proscript}.

For our algorithm \ours\, we omitted the event expansion stage since the event skeleton construction stage already generated enough events.
In the event-event relation verification stage, we continue to add temporal relations among events based on their verification score beyond the threshold until the graph is connected. 

\begin{table}[t]
\setlength{\tabcolsep}{3.5pt}
    \centering
    \small 
    \begin{tabular}{l|c c a c c a}
    \toprule 
        &  \multicolumn{3}{c}{Event} & \multicolumn{3}{c}{Temp. Relation}    \\
        Model & Prec & Recall & F1 & Prec & Recall &F1 \\
        \midrule 
        GPT3-DOT & 61.8 & 59.5 & 59.3 & 25.9& 23.3 & \textbf{23.7} \\
        \ours & 58.4 & 69.6 & \textbf{61.1} & 22.4 & 25.8 & 22.7\\
        \bottomrule 
    \end{tabular}
    \caption{Evaluation on Proscript's everyday scenarios. These schemas are small in size and do not contain hierarchical relations.  We compare against directly generating the linearized graph as DOT language.
    }
    \label{tab:exp-proscript}
\end{table}

On these small-scaled schemas with only temporal relations, we see that directly generating the full schema and incrementally prompting the schema lead to comparable results. This shows that GPT3 can indeed understand and generate valid graph description DOT language and the gap that we observe in Table \ref{tab:exp-phase2b} is mainly due to the difficulty of capturing long-range dependencies in large schemas and the confusion between temporal and hierarchical relations.

%% file: 5_related.tex
\section{Related Work}
\paragraph{Event Schema Induction}
Event schema induction, or script induction, is the task of inducing typical event-event relation structures for given scenarios/situations\footnote{There exists some work that refer to the task of ``inducing roles of events'' as schema induction, but their scope is distinct from ours.}. A large fraction of work considers event schemas as narrative chains~\cite{chambers-jurafsky-2008-unsupervised,chambers-jurafsky-2009-unsupervised, jans-etal-2012-skip,pichotta-mooney-2014-statistical,pichotta-mooney-2016-statistical, rudinger-etal-2015-script, ahrendt-demberg-2016-improving,GranrothWilding2016WhatHN, wang-etal-2017-integrating,Weber2018EventRW}, limiting the structure to include only sequential temporal relations. More recently, non-sequential, partially ordered temporal relations have been taken into consideration~\cite{Li2018ConstructingNarrativeGraph,li-etal-2020-connecting,li-etal-2021-future,sakaguchi-etal-2021-proscript-partially, jin-etal-2022-event} but they do not consider the different scales of events and the potential hierarchical relations. In terms of schema expressiveness, \cite{dror2022zeroshotschema} is the most similar to ours as they also consider both partial temporal order and hierarchical relations. 

Our work also resembles a line of recent research on inducing schema knowledge from pre-trained language models.
Our schema induction process can be seen as a super-set of the post-processing in \cite{sancheti-rudinger-2022-large}, which comprises irrelevant event removal, de-duplication, and temporal relation correction. We compare our incremental prompting approach with the end-to-end approach proposed in \cite{sakaguchi-etal-2021-proscript-partially} in Section \ref{sec:exp}. The work of \cite{dror2022zeroshotschema} is orthogonal to ours as they use LLMs for data generation instead of probing for schema knowledge.

\paragraph{Language Model Prompting}
Prompting has been the major method of interaction with billion-scale language models~\cite{Brown2020GPT3,  Rae2021Gopher, Wei2021FLAN, Chowdhery2022PaLM}.
Prompting can either be used to inform the model of the task instructions~\cite{Wei2021FLAN}, provide the model with task input/output examples~\cite{Brown2020GPT3}, or guide the model with explanations~\cite{Lampinen2022CanLM} and reasoning paths~\cite{Wang2022RationaleAugmentedEI}. 
In this work, we explore how a complex knowledge structure such as an event graph schema can be induced using LLMs by decomposing the task through incremental prompting.

%% file: 6_conclusion.tex
\section{Conclusions and Future Work}
Prior work on schema induction has either relied on existing information extraction pipelines to convert unstructured documents into event graphs, or require massive human effort in annotating event schemas. 
We propose to view schema induction as a type of \textit{event-oriented commonsense} that can be implicitly learned with large language models. 
However, since schemas are complex graph structures, instead of directly querying for schemas, we design an incremental prompting and verification framework \ours\ to decompose the schema induction task into a series of simple questions. 
As a result, our model is applicable to the open-domain and can jointly induce temporal and hierarchical relations between events.

For future work, we plan to cover more aspects of schemas, including accounting for entity coreference, entity relations and entity attributes. 
While this work is focused on the task of schema induction, we hope to show the possibility of using LLMs for constructing complex knowledge structures.

%% file: 7_limitations.tex
\section{Limitations}
The event schemas generated by our model are not directly comparable to those generated by previous work that utilized a close-domain ontology. As a result, we were unable to adopt the same metrics and evaluate our schemas on type-level event prediction tasks as in \cite{li-etal-2021-future, jin-etal-2022-event,dror2022zeroshotschema}. Grounding the events generated by the LLM into one of the types in the ontology could be added as a post-processing step to our model, but this would require some ontology-specific training data, which goes against our principles of designing an \textit{open-domain, portable} framework.

Our event schema does not explicitly represent entity coreference, entity relations, and entity attributes. The current schemas that we produce focus on events and their relations, with entity information captured implicitly through the event descriptions. 
For instance, the \texttt{See Medical Professional} event is described as ``The patient is seen by a doctor or other medical professional'' and the proceeding \texttt{Obtain Medical History} event is described as ``The medical professional obtains a medical history from the patient''. The ``medical professional'' and ``patient'' are implied to be coreferential entities in this case, but not explicitly connected in the schema graph.

Our approach is also quite distinct from prior work~\cite{rudinger2015script,wang-etal-2017-integrating,li-etal-2021-future, jin-etal-2022-event} that consider a probabilistic model as an implicit schema where the schema graph, or event narrative chain can be sampled from. Probabilistic schema models have the advantage of being adaptive and can be conditioned on partially observed event sequences, but are hard to interpret. We make the conscious design decision to generate explicit, human-readable schema graphs instead of black-box schema models. 

Finally, our model relies on the usage of LMs, which have been observed to sometimes show inconsistent behavior between different runs or when using different prompts with the same meaning~\cite{elazar-etal-2021-measuring,Zhou2022PromptConsistency}. However, quantification of consistency has only been done for factual probing tasks while schema generation is a more open-ended task. For example, in our experiments on everyday scenarios, we observe that the model could generate distinct schemas for \texttt{Buying a (computer) mouse} based on whether the purchase was done online or in person. 
This variance is often benign and we leave it to future work to take advantage of such variance and possibly aggregate results over multiple runs.

%% file: appendix.tex
\section{Retrieval Component}
\label{sec:appendix_retrieval} 
To build our document collection, we first search for the scenario name on Wikipedia, find its corresponding category page\footnote{A category page is a curated list of Wikipedia pages organized by topic, such as \url{https://en.wikipedia.org/wiki/Category:Disease_outbreaks}.} and then for each Wikipedia article listed under the category, we follow the external reference links to news sources under the Wikipedia pages to retrieve the original news articles. 
We only keep English articles and filter out articles that have fewer than 4 sentences. Then we split the articles into overlapping segments of 5 sentences with 1 sentence overlap for indexing. 

Our retrieval model is based on 
TCT-ColBERT~\cite{lin-etal-2021-batch}, specifically, the implementation provided by Pyserini~\cite{Lin_etal_SIGIR2021_Pyserini} and pretrained on the MSMARCO dataset~\cite{Campos2016MSMARCO}. TCT-ColBERT is a distillation of ColBERT~\cite{Khattab2020ColBERT} which is a late-interaction bi-encoder model. 
It encodes the query and the document separately into multiple vectors offline and then employs an interaction step to compute their similarity.

\section{Algorithm for Removing Temporal Loops}
\label{sec:appendix_a}
The key observation for finding the minimum feedback arc set is to convert the problem into finding an ordering $v_1 v_2 \cdots v_n$ among the vertices of graph $G$, then all of the edges $v_i v_j$ that violate this ordering by having $i>j$ will be feedback arcs.

To create a good ordering (with a small number of feedback arcs), we maintain two lists $s_1$ and $s_2$ which correspond to the head and tail of the vertex ordering. 
We first remove the source and sink nodes from the graph recursively by adding the source nodes to $s_1$ and the sink nodes to $s_2$.

For the remaining nodes, we compute a $\delta(v)$ score for each node which is the difference between the weights of its outgoing edges and incoming edges. Then the node with the maximal $\delta(v)$ will be appended to the end of $s_1$ and removed from the graph.
This step is also done recursively and $\delta$ needs to be recomputed after removing nodes. 
Finally the ordering is obtained by concatenating $s_1$ and $s_2$. 
The complete algorithm is shown in Algorithm \ref{alg:fas}.

This ordering will divide the edges in graph $G$ into 2 sets: the set of edges $(v_i, v_j)$ that follow the ordering $i<j$ and the set of edges that go against the ordering $j>i$. The feedback arc set will be whichever of these two sets that have lesser edges. 

\begin{algorithm}[ht]
\caption{A greedy algorithm for finding the minimal feedback arc set~\cite{eades1993fast} }
\label{alg:fas}
\begin{algorithmic}
\Require Graph $G$
\State $s_1 \leftarrow \emptyset$, $s_2 \leftarrow \emptyset$ 
\While {$G \neq \emptyset$}
\While {$G$ contains sink node $v$}
\State $G \leftarrow G - v$
\State $s_2 \leftarrow vs_2$
\EndWhile 
\While {$G$ contains source node $v$}
\State $G \leftarrow G - v$
\State $s_1 \leftarrow s_1v$
\EndWhile 
\State $v = \arg\max_G \delta(v)$
\State $G \leftarrow G - v$
\EndWhile
\State $s \leftarrow s_1 s_2$
\end{algorithmic}
\end{algorithm}

\section{List of Scenario Names}
\label{sec:appendix_scenarios}
We show the complete list of scenarios in RESIN-11 and \dataset\ in Table \ref{tab:scenarios}.

All of our scenario documents and schemas are in English.

\begin{table}[t]
    \centering
    \small 
    \begin{tabular}{c|l}
    \toprule 
    Dataset &  Scenarios \\
    \midrule 
      RESIN-11   &  Business change \\
      & Election \\
      & General IED \\
      & Kidnapping \\
      & Mass shooting \\
      & Natural disaster and rescue \\
      & Sports \\
      & Disease outbreak \\
      & Civil unrest (Protest) \\
      & Terrorist attack \\
      & International conflict \\
      \midrule 
    \dataset & Chemical spill \\
      & Chemical warfare \\
      & Coup \\
      & Cyber attack \\
      & Health care \\
      & Infrastructure disaster \\
      & International aggression \\
      & Investment \\
      & Medical procedure \\
      & Medical research \\
      & Nuclear attack \\
      & Political corruption \\
      & Recession \\
      & Refugee crisis \\
      & Trading \\
      & Transport accident \\
      & Violent crime \\
      & Warfare \\
      \bottomrule 
    \end{tabular}
    \caption{The complete list of scenarios that were used in our experiments. RESIN-11 provides many variants of the IED scenario, we kept the \texttt{General IED} scenario.}
    \label{tab:scenarios}
\end{table}

\section{List of Prompts and In-Context Examples}
\label{sec:appendix_prompt}

\subsection{Templated Prompts in \ours}
Below is the prompt that we use for the \textit{event skeleton construction} stage: 
\begin{lstlisting}{text}
{evt.name} is defined as "{evt.description}". 
List the major events that happen in the {evt.name} of a {scenario}:
\end{lstlisting}

\texttt{scenario} is the scenario name and 
\texttt{evt} can be filled in with the chapters that are provided as part of the input.

To assign names to the events, we use the following prompt with 10 in-context examples: 
\begin{lstlisting}{text}
Give names to the described event. 
Description: Disinfect the area to prevent infection of the disease. Name: Sanitize 
Description: A viral test checks specimens from your nose or your mouth to find out if you are currently infected with the virus. Name: Test for Virus  
Description: If the jury finds the defendant guilty, they may be sentenced to jail time, probation, or other penalties. Name: Sentence
Description: The police or other law enforcement officials arrive at the scene of the bombing. Name: Arrive at Scene
Description: The attacker parks the vehicle in a location that will cause maximum damage and casualties.  Name: Park Vehicle
Description: The government declares a state of emergency. Name: Declare Emergency
Description: The government mobilizes resources to respond to the outbreak. Name: Mobilize Resources
Description: The liable party is required to pay damages to the affected parties. Name: Pay Damages
Description: People declare candidacy and involve in the campaign for party nomination. Name: Declare Candidacy 
Description: Assessing the damage caused by the disaster and working on a plan to rebuild. Name: Assess Damage
Description:{evt.description} Name: 
\end{lstlisting}

For the second \textit{event expansion} stage, we use the following 6 prompts: 

\begin{lstlisting}{text}
What happened during "{evt.description}"? List the answers: 
What are the steps in "{evt.description}"? List the answers: 
What happens after "{evt.description}"? List the answers: 
What happened before "{evt.description}"? List the answers: 
List the consequences of "{evt.description}:
List the possible causes of "{evt.description}":
\end{lstlisting}

The prompt for the specificity test containing 10 in-context examples is: 
\begin{lstlisting}{text}
Does the text contain any specific names, numbers, locations or dates? Answer yes or no. 

Text: The UN Strategy for Recovery is launched in an attempt to rebuild the areas most affected by the Chernobyl disaster. Answer: Yes
Text: More than 300 teachers in the Jefferson County school system took advantage of counseling services. Answer: Yes 
Text: The police or other law enforcement officials will interview witnesses and potential suspects. Answer: No
Text: The IHT will establish a Defense Office to ensure adequate facilities for counsel in the preparation of defense cases. Answer: Yes
Text: Helping people to recover emotionally and mentally from the trauma of the disaster. Answer: No
Text: The area is cleaned up and any contaminated materials are removed. Answer: No
Text: About 100,000 people evacuated Mariupol. Answer: Yes
Text: Gabriel Aduda said three jets chartered from local carriers would leave the country on Wednesday. Answer: Yes
Text: The party attempting the coup becomes increasingly frustrated with the ruling government. Answer: No
Text: The international community condemns the war and calls for a peaceful resolution: Answer: No
Text: {evt.description} Answer: 
\end{lstlisting}

Examples of events that \textbf{did not} pass the specificity test: 
\begin{lstlisting}{text}
Reporting of the first suspected cases in the limits of Union Council 39 of Tehkal Bala area
Focus groups with members of the public from 5 provinces were conducted to identify major factors influencing public knowledge, perceptions and behaviours during COVID
The PLO sent an encrypted message to the Iraqi Foreign Ministry in Baghdad
\end{lstlisting}

The prompt for the chapter test is 
\begin{lstlisting}{text}
{chapter_evt.name} is defined as "{chapter_evt.description}"
{evt.name} is defined as "{evt.description}" 
Is {evt.name} a part of {chapter_evt.name}? Answer yes or no. 
\end{lstlisting}

For the \textit{event-event relation verification} stage, we use the following questions:
\begin{lstlisting}{text}
Does "{e1.description}" start before "{e2.description}"? Answer yes, no or unknown.
Does "{e1.description}" end before "{e2.description}"? Answer yes, no or unknown.
Is the duration of {e1.description} longer than {e2.description}? Answer yes or no.
\end{lstlisting}

Note that in the verification stage, we use the probabilities assigned to the ``yes'', ``no'', and ``unknown'' tokens instead of directly taking the generated text as the answer.

\subsection{In-Context Example for GPT3-DOT}
The in-context example follows the DOT language specifications (\url{https://graphviz.org/doc/info/lang.html}) to linearize a graph. Here we only list a few events and relations due to length considerations.

\begin{lstlisting}{text}
List relevant events and edges in "chemical spills":
events:
0: chemical spills news story.
1: The accumulation of the chemical that is leaked later.
2: Chemicals are spilled into the environment.
3: The spill causes other hazards such as fire.
...
edges: 
1->2[label='temporal']
2->3[label='temporal']
3->4[label='temporal']
...
\end{lstlisting}

\section{Schema Examples}
\begin{figure*}[ht]
    \centering
    \includegraphics[width=.8\linewidth]{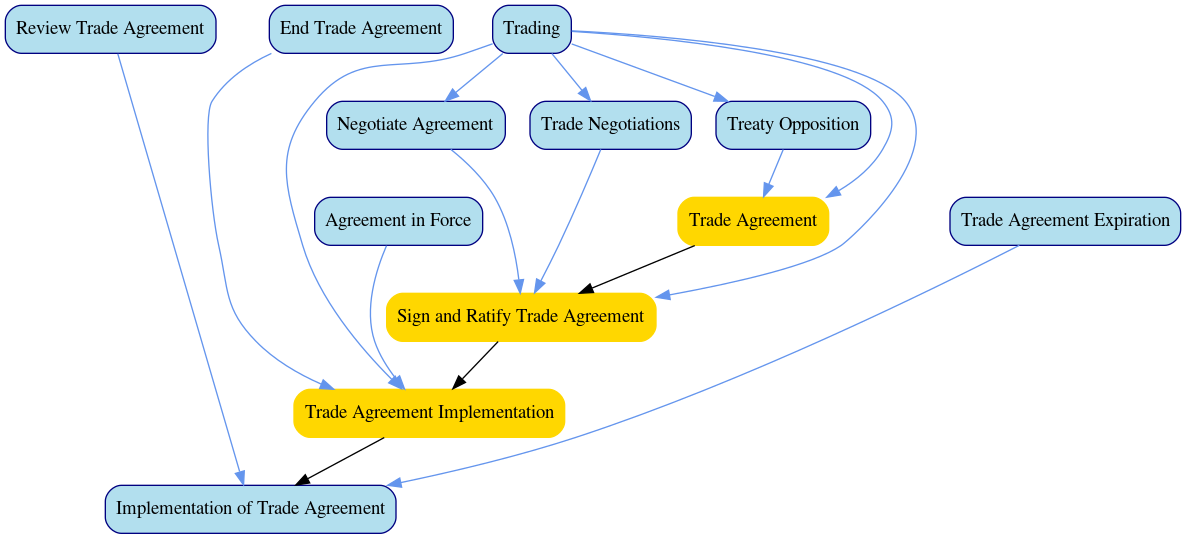}
    \caption{The Trading schema generated by GPT3-DOT.}
    \label{fig:gptdot_output}
\end{figure*}

\begin{figure*}
    \centering
    \includegraphics[width=\linewidth]{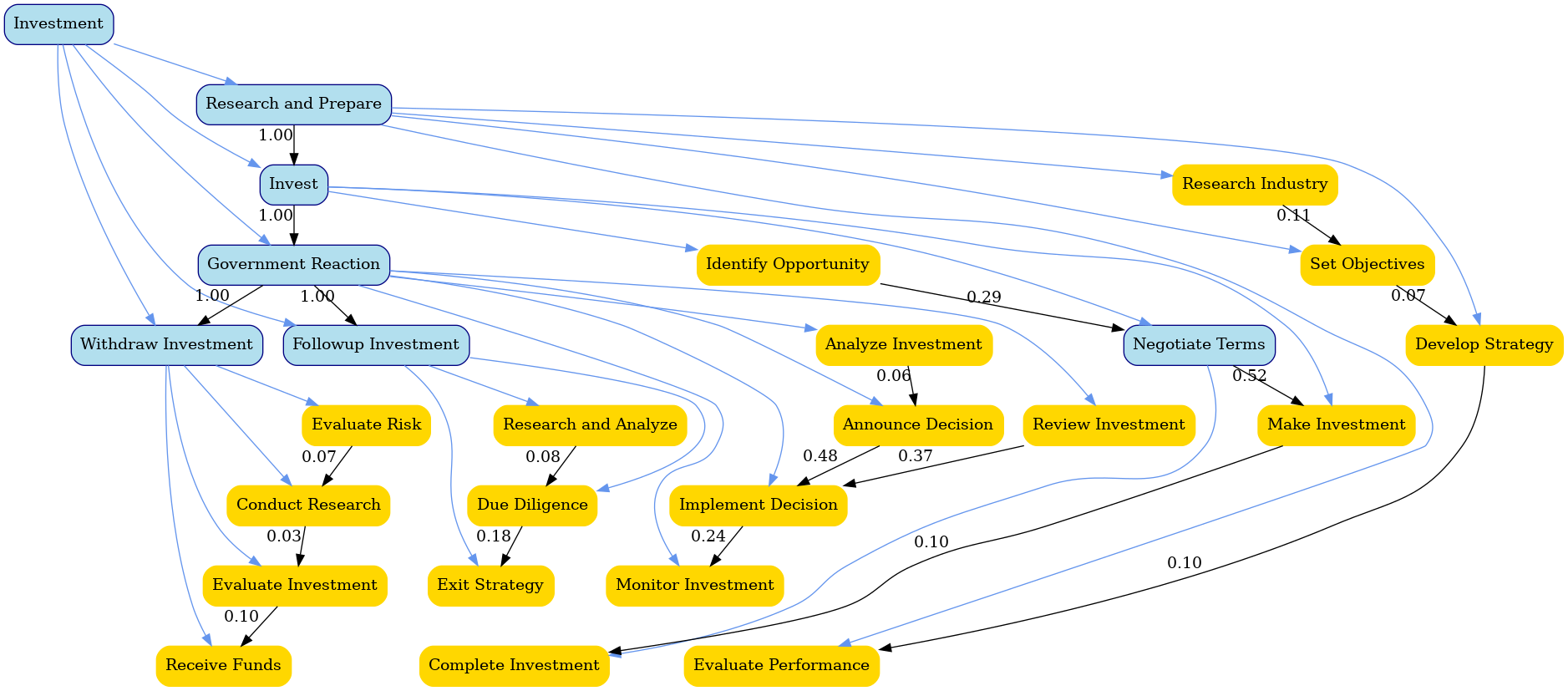}
    \caption{The schema for Investment generated by \ours.}
    \label{fig:investment_schema}
\end{figure*}

We show an example of a schema generated by GPT3-DOT in Figure \ref{fig:gptdot_output} and an example schema generated by \ours\ in \ref{fig:investment_schema}. 
In the visualization, blue nodes are events with subevents (children nodes) and yellow nodes are primitive events (leaf nodes). 
Blue edges represent hierarchical relations and go from parent to child. 
Black edges represent temporal edges and go from the previous event to the proceeding event. 
Schemas generated by GPT3-DOT are typically much smaller in size and confuse hierarchical relations with temporal relations.

\section{Human Assessment Details} 
\label{sec:appendix_interface} 
\begin{figure*}[ht]
    \centering
    \includegraphics[width=.8\linewidth]{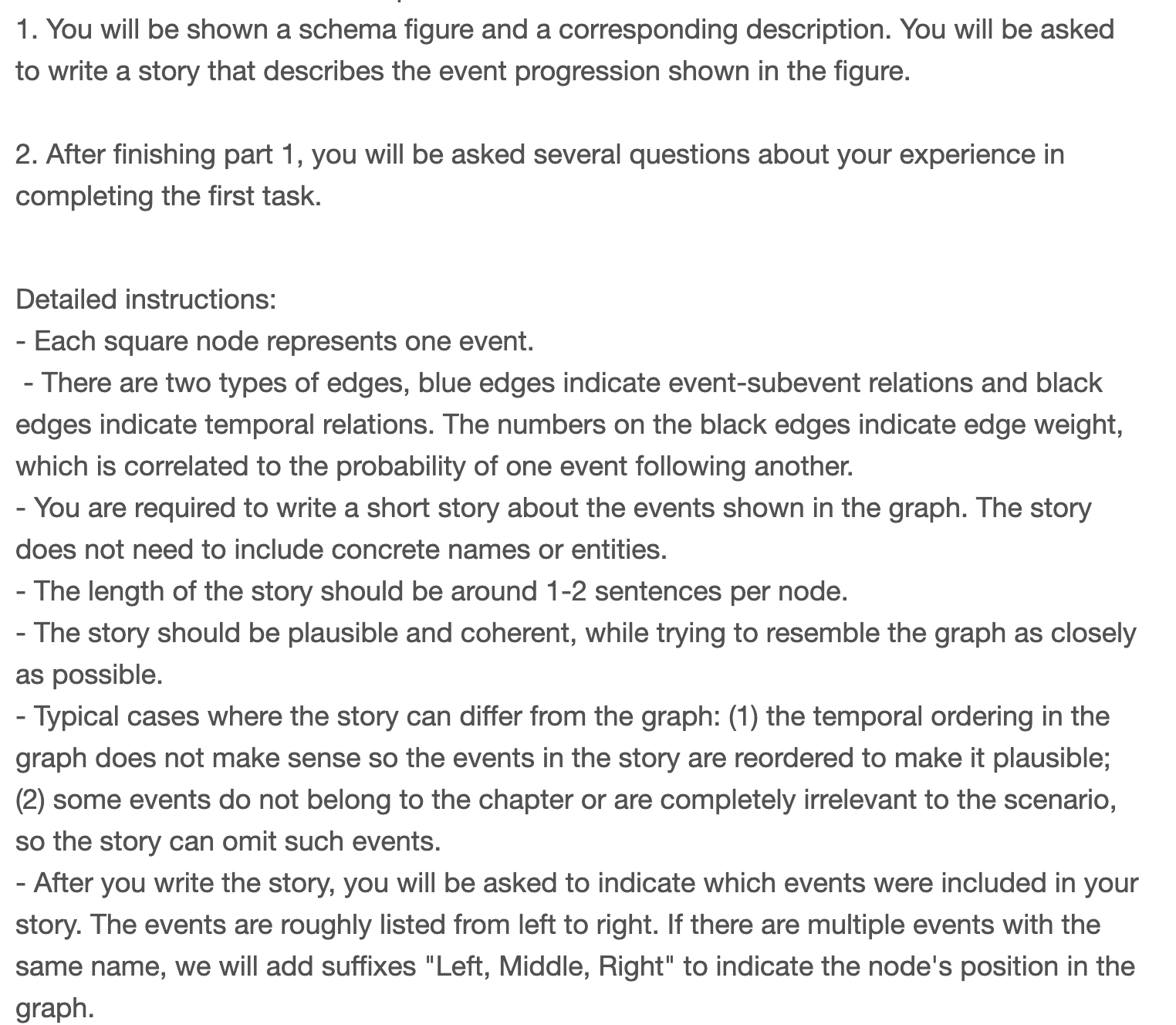}
    \caption{Full set of instructions shown to assessors.}
    \label{fig:instructions}
\end{figure*}

\begin{figure*}[ht]
    \centering
    \includegraphics[width=\linewidth]{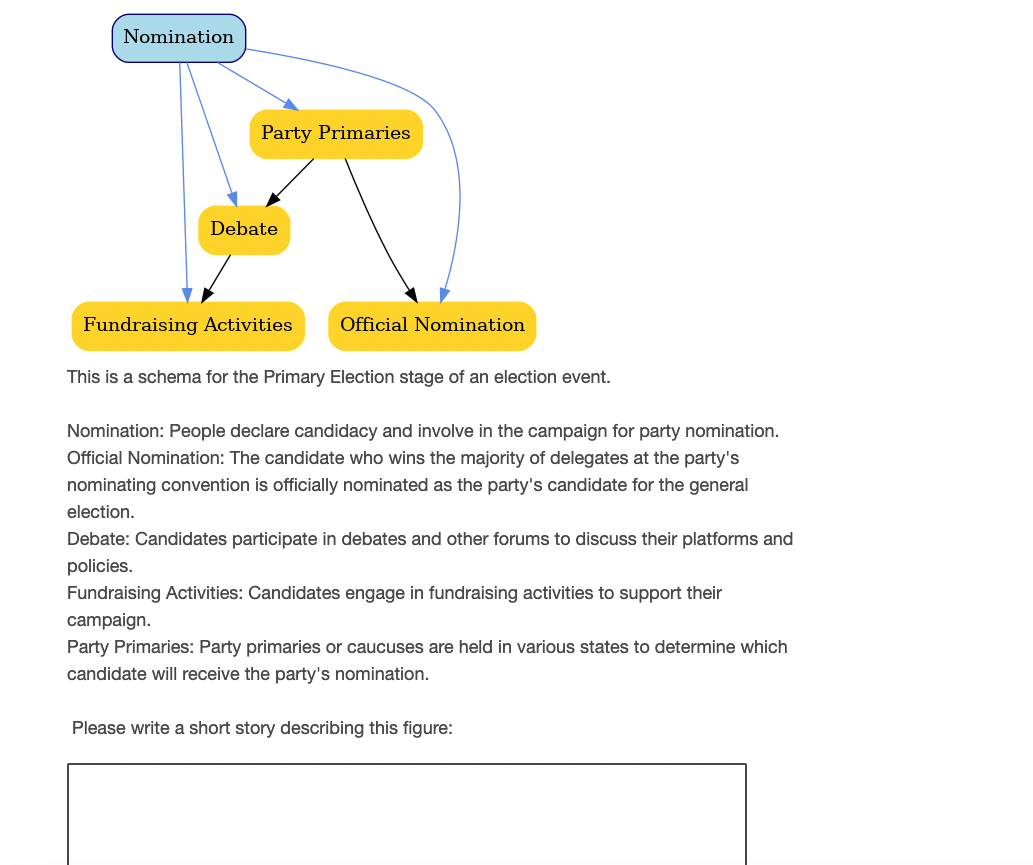}
    \caption{The interface for human assessment. The assessor is shown a figure of the schema and descriptions of the events (if applicable).}
    \label{fig:interface}
\end{figure*}

\begin{figure*}[ht]
    \centering
    \includegraphics[width=.6\linewidth]{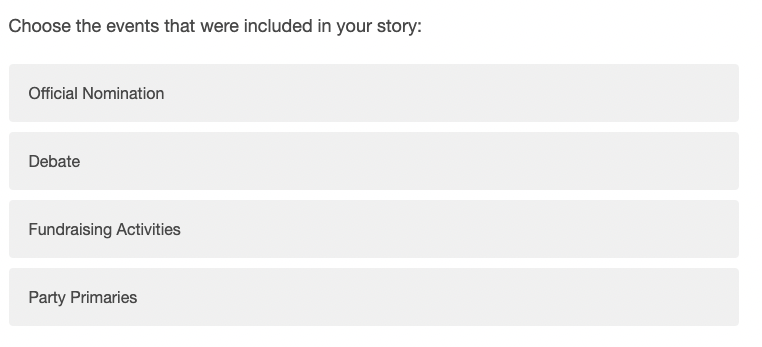}
    \caption{After the assessors write the story for the schema, they will be asked to choose which events were included in the story.}
    \label{fig:interface2}
\end{figure*}

\begin{figure*}[ht]
    \centering
    \includegraphics[width=\linewidth]{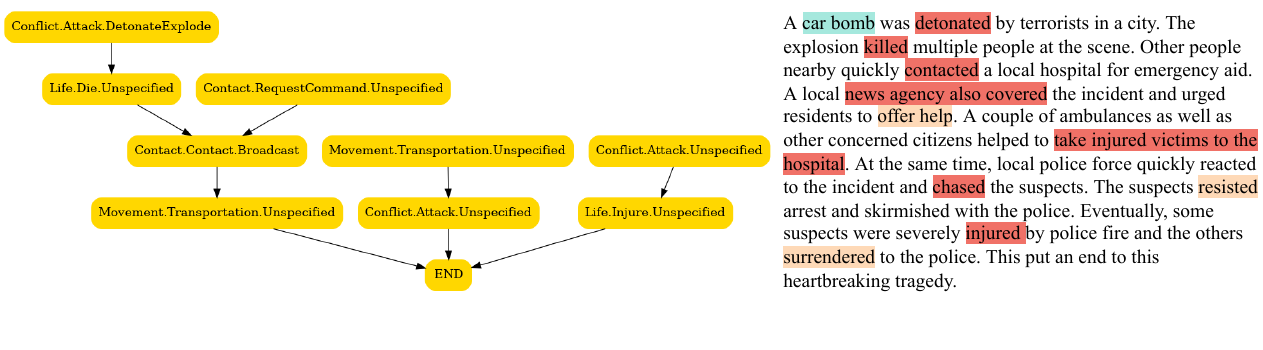}
    \caption{An example schema from Double-GAE~\cite{jin-etal-2022-event} and human response for the interpretability assessment.}
    \label{fig:human_assessment_DGAE}
\end{figure*}

We designed and distributed our human assessment task with Qualtrics\footnote{\url{https://www.qualtrics.com/}}.
We recruited 15 graduate students as our human assessors (all of which are paid as research assistants).
The assessors had basic knowledge of what a schema is, but were not involved in the development of our model. Assessors were informed of the purpose of the study.
Before they begin to work on the story-writing task, they were presented with task instructions (Figure \ref{fig:instructions} and an example response. We did not collect any personal identifiers during the assessment. 
The order of the schema graphs is randomized both in terms of the schema induction algorithm and the scenario. We show two screenshots of the interface in Figure \ref{fig:interface} and Figure \ref{fig:interface2}.
Additionally, we show a figure of the schema generated by Double-GAE and a human response corresponding to the schema in Figure \ref{fig:human_assessment_DGAE}.

%% file: acl_latex.bbl
\begin{thebibliography}{42}
\expandafter\ifx\csname natexlab\endcsname\relax\def\natexlab#1{#1}\fi

\bibitem[{Ahrendt and Demberg(2016)}]{ahrendt-demberg-2016-improving}
Simon Ahrendt and Vera Demberg. 2016.
\newblock \href {https://doi.org/10.18653/v1/N16-1067} {Improving event
  prediction by representing script participants}.
\newblock In \emph{Proceedings of the 2016 Conference of the North {A}merican
  Chapter of the Association for Computational Linguistics: Human Language
  Technologies}, pages 546--551, San Diego, California. Association for
  Computational Linguistics.

\bibitem[{Allen(1983)}]{Allen1983MaintainingKA}
James~F. Allen. 1983.
\newblock Maintaining knowledge about temporal intervals.
\newblock \emph{Commun. ACM}, 26:832--843.

\bibitem[{Brown et~al.(2020)Brown, Mann, Ryder, Subbiah, Kaplan, Dhariwal,
  Neelakantan, Shyam, Sastry, Askell, Agarwal, Herbert-Voss, Krueger, Henighan,
  Child, Ramesh, Ziegler, Wu, Winter, Hesse, Chen, Sigler, Litwin, Gray, Chess,
  Clark, Berner, McCandlish, Radford, Sutskever, and Amodei}]{Brown2020GPT3}
Tom~B. Brown, Benjamin Mann, Nick Ryder, Melanie Subbiah, Jared Kaplan,
  Prafulla Dhariwal, Arvind Neelakantan, Pranav Shyam, Girish Sastry, Amanda
  Askell, Sandhini Agarwal, Ariel Herbert-Voss, Gretchen Krueger, T.~J.
  Henighan, Rewon Child, Aditya Ramesh, Daniel~M. Ziegler, Jeff Wu, Clemens
  Winter, Christopher Hesse, Mark Chen, Eric Sigler, Mateusz Litwin, Scott
  Gray, Benjamin Chess, Jack Clark, Christopher Berner, Sam McCandlish, Alec
  Radford, Ilya Sutskever, and Dario Amodei. 2020.
\newblock Language models are few-shot learners.
\newblock \emph{Neurips}.

\bibitem[{Campos et~al.(2016)Campos, Nguyen, Rosenberg, Song, Gao, Tiwary,
  Majumder, Deng, and Mitra}]{Campos2016MSMARCO}
Daniel~Fernando Campos, Tri Nguyen, Mir Rosenberg, Xia Song, Jianfeng Gao,
  Saurabh Tiwary, Rangan Majumder, Li~Deng, and Bhaskar Mitra. 2016.
\newblock Ms marco: A human generated machine reading comprehension dataset.
\newblock \emph{ArXiv}, abs/1611.09268.

\bibitem[{Chambers and Jurafsky(2008)}]{chambers-jurafsky-2008-unsupervised}
Nathanael Chambers and Dan Jurafsky. 2008.
\newblock \href {https://aclanthology.org/P08-1090} {Unsupervised learning of
  narrative event chains}.
\newblock In \emph{Proceedings of ACL-08: HLT}, pages 789--797, Columbus, Ohio.
  Association for Computational Linguistics.

\bibitem[{Chambers and Jurafsky(2009)}]{chambers-jurafsky-2009-unsupervised}
Nathanael Chambers and Dan Jurafsky. 2009.
\newblock \href {https://aclanthology.org/P09-1068} {Unsupervised learning of
  narrative schemas and their participants}.
\newblock In \emph{Proceedings of the Joint Conference of the 47th Annual
  Meeting of the {ACL} and the 4th International Joint Conference on Natural
  Language Processing of the {AFNLP}}, pages 602--610, Suntec, Singapore.
  Association for Computational Linguistics.

\bibitem[{Chowdhery et~al.(2022)Chowdhery, Narang, Devlin, Bosma, Mishra,
  Roberts, Barham, Chung, Sutton, Gehrmann, Schuh, Shi, Tsvyashchenko, Maynez,
  Rao, Barnes, Tay, Shazeer, Prabhakaran, Reif, Du, Hutchinson, Pope, Bradbury,
  Austin, Isard, Gur-Ari, Yin, Duke, Levskaya, Ghemawat, Dev, Michalewski,
  Garc{\'i}a, Misra, Robinson, Fedus, Zhou, Ippolito, Luan, Lim, Zoph,
  Spiridonov, Sepassi, Dohan, Agrawal, Omernick, Dai, Pillai, Pellat,
  Lewkowycz, Moreira, Child, Polozov, Lee, Zhou, Wang, Saeta, D{\'i}az, Firat,
  Catasta, Wei, Meier-Hellstern, Eck, Dean, Petrov, and
  Fiedel}]{Chowdhery2022PaLM}
Aakanksha Chowdhery, Sharan Narang, Jacob Devlin, Maarten Bosma, Gaurav Mishra,
  Adam Roberts, Paul Barham, Hyung~Won Chung, Charles Sutton, Sebastian
  Gehrmann, Parker Schuh, Kensen Shi, Sasha Tsvyashchenko, Joshua Maynez,
  Abhishek~B Rao, Parker Barnes, Yi~Tay, Noam~M. Shazeer, Vinodkumar
  Prabhakaran, Emily Reif, Nan Du, Benton~C. Hutchinson, Reiner Pope, James
  Bradbury, Jacob Austin, Michael Isard, Guy Gur-Ari, Pengcheng Yin, Toju Duke,
  Anselm Levskaya, Sanjay Ghemawat, Sunipa Dev, Henryk Michalewski, Xavier
  Garc{\'i}a, Vedant Misra, Kevin Robinson, Liam Fedus, Denny Zhou, Daphne
  Ippolito, David Luan, Hyeontaek Lim, Barret Zoph, Alexander Spiridonov, Ryan
  Sepassi, David Dohan, Shivani Agrawal, Mark Omernick, Andrew~M. Dai,
  Thanumalayan~Sankaranarayana Pillai, Marie Pellat, Aitor Lewkowycz,
  Erica~Oliveira Moreira, Rewon Child, Oleksandr Polozov, Katherine Lee,
  Zongwei Zhou, Xuezhi Wang, Brennan Saeta, Mark D{\'i}az, Orhan Firat, Michele
  Catasta, Jason Wei, Kathleen~S. Meier-Hellstern, Douglas Eck, Jeff Dean, Slav
  Petrov, and Noah Fiedel. 2022.
\newblock Palm: Scaling language modeling with pathways.
\newblock \emph{ArXiv}, abs/2204.02311.

\bibitem[{Dror et~al.(2022)Dror, Wang, and Roth}]{dror2022zeroshotschema}
Rotem Dror, Haoyu Wang, and Dan Roth. 2022.
\newblock \href {https://doi.org/10.48550/ARXIV.2210.06254} {Zero-shot
  on-the-fly event schema induction}.

\bibitem[{Du et~al.(2022)Du, Zhang, Li, Yu, Wang, Lai, Lin, Wang, Liu, Zhou,
  Wen, Li, Hannan, Lei, Kim, Dror, Wang, Regan, Zeng, Lyu, Yu, Edwards, Jin,
  Jiao, Kazeminejad, Wang, Callison-Burch, Bansal, Vondrick, Han, Roth, Chang,
  Palmer, and Ji}]{du-etal-2022-resin}
Xinya Du, Zixuan Zhang, Sha Li, Pengfei Yu, Hongwei Wang, Tuan Lai, Xudong Lin,
  Ziqi Wang, Iris Liu, Ben Zhou, Haoyang Wen, Manling Li, Darryl Hannan, Jie
  Lei, Hyounghun Kim, Rotem Dror, Haoyu Wang, Michael Regan, Qi~Zeng, Qing Lyu,
  Charles Yu, Carl Edwards, Xiaomeng Jin, Yizhu Jiao, Ghazaleh Kazeminejad,
  Zhenhailong Wang, Chris Callison-Burch, Mohit Bansal, Carl Vondrick, Jiawei
  Han, Dan Roth, Shih-Fu Chang, Martha Palmer, and Heng Ji. 2022.
\newblock \href {https://doi.org/10.18653/v1/2022.naacl-demo.7} {{RESIN}-11:
  Schema-guided event prediction for 11 newsworthy scenarios}.
\newblock In \emph{Proceedings of the 2022 Conference of the North American
  Chapter of the Association for Computational Linguistics: Human Language
  Technologies: System Demonstrations}, pages 54--63, Hybrid: Seattle,
  Washington + Online. Association for Computational Linguistics.

\bibitem[{Eades et~al.(1993)Eades, Lin, and Smyth}]{eades1993fast}
Peter Eades, Xuemin Lin, and William~F Smyth. 1993.
\newblock A fast and effective heuristic for the feedback arc set problem.
\newblock \emph{Information Processing Letters}, 47(6):319--323.

\bibitem[{Elazar et~al.(2021)Elazar, Kassner, Ravfogel, Ravichander, Hovy,
  Sch{\"u}tze, and Goldberg}]{elazar-etal-2021-measuring}
Yanai Elazar, Nora Kassner, Shauli Ravfogel, Abhilasha Ravichander, Eduard
  Hovy, Hinrich Sch{\"u}tze, and Yoav Goldberg. 2021.
\newblock \href {https://doi.org/10.1162/tacl_a_00410} {Measuring and improving
  consistency in pretrained language models}.
\newblock \emph{Transactions of the Association for Computational Linguistics},
  9:1012--1031.

\bibitem[{Glava{\v{s}} et~al.(2014)Glava{\v{s}}, {\v{S}}najder, Moens, and
  Kordjamshidi}]{glavas-etal-2014-hieve}
Goran Glava{\v{s}}, Jan {\v{S}}najder, Marie-Francine Moens, and Parisa
  Kordjamshidi. 2014.
\newblock \href
  {http://www.lrec-conf.org/proceedings/lrec2014/pdf/1023_Paper.pdf}
  {{H}i{E}ve: A corpus for extracting event hierarchies from news stories}.
\newblock In \emph{Proceedings of the Ninth International Conference on
  Language Resources and Evaluation ({LREC}'14)}, pages 3678--3683, Reykjavik,
  Iceland. European Language Resources Association (ELRA).

\bibitem[{Granroth-Wilding and Clark(2016)}]{GranrothWilding2016WhatHN}
Mark Granroth-Wilding and Stephen Clark. 2016.
\newblock What happens next? event prediction using a compositional neural
  network model.
\newblock In \emph{AAAI}.

\bibitem[{Jans et~al.(2012)Jans, Bethard, Vuli{\'c}, and
  Moens}]{jans-etal-2012-skip}
Bram Jans, Steven Bethard, Ivan Vuli{\'c}, and Marie~Francine Moens. 2012.
\newblock \href {https://aclanthology.org/E12-1034} {Skip n-grams and ranking
  functions for predicting script events}.
\newblock In \emph{Proceedings of the 13th Conference of the {E}uropean Chapter
  of the Association for Computational Linguistics}, pages 336--344, Avignon,
  France. Association for Computational Linguistics.

\bibitem[{Ji and Grishman(2008)}]{ji2008refining}
Heng Ji and Ralph Grishman. 2008.
\newblock Refining event extraction through unsupervised cross-document
  inference.
\newblock In \emph{In Proceedings of the Annual Meeting of the Association of
  Computational Linguistics (ACL 2008). Ohio, USA}.

\bibitem[{Jin et~al.(2022)Jin, Li, and Ji}]{jin-etal-2022-event}
Xiaomeng Jin, Manling Li, and Heng Ji. 2022.
\newblock \href {https://doi.org/10.18653/v1/2022.naacl-main.147} {Event schema
  induction with double graph autoencoders}.
\newblock In \emph{Proceedings of the 2022 Conference of the North American
  Chapter of the Association for Computational Linguistics: Human Language
  Technologies}, pages 2013--2025, Seattle, United States. Association for
  Computational Linguistics.

\bibitem[{Khattab and Zaharia(2020)}]{Khattab2020ColBERT}
O.~Khattab and Matei~A. Zaharia. 2020.
\newblock Colbert: Efficient and effective passage search via contextualized
  late interaction over bert.
\newblock \emph{Proceedings of the 43rd International ACM SIGIR Conference on
  Research and Development in Information Retrieval}.

\bibitem[{Lampinen et~al.(2022)Lampinen, Dasgupta, Chan, Matthewson, Tessler,
  Creswell, McClelland, Wang, and Hill}]{Lampinen2022CanLM}
Andrew~Kyle Lampinen, Ishita Dasgupta, Stephanie C.~Y. Chan, Kory Matthewson,
  Michael~Henry Tessler, Antonia Creswell, James~L. McClelland, Jane~X. Wang,
  and Felix Hill. 2022.
\newblock Can language models learn from explanations in context?
\newblock \emph{ArXiv}, abs/2204.02329.

\bibitem[{Li et~al.(2021)Li, Li, Wang, Huang, Cho, Ji, Han, and
  Voss}]{li-etal-2021-future}
Manling Li, Sha Li, Zhenhailong Wang, Lifu Huang, Kyunghyun Cho, Heng Ji,
  Jiawei Han, and Clare Voss. 2021.
\newblock \href {https://doi.org/10.18653/v1/2021.emnlp-main.422} {The future
  is not one-dimensional: Complex event schema induction by graph modeling for
  event prediction}.
\newblock In \emph{Proceedings of the 2021 Conference on Empirical Methods in
  Natural Language Processing}, pages 5203--5215, Online and Punta Cana,
  Dominican Republic. Association for Computational Linguistics.

\bibitem[{Li et~al.(2020)Li, Zeng, Lin, Cho, Ji, May, Chambers, and
  Voss}]{li-etal-2020-connecting}
Manling Li, Qi~Zeng, Ying Lin, Kyunghyun Cho, Heng Ji, Jonathan May, Nathanael
  Chambers, and Clare Voss. 2020.
\newblock \href {https://doi.org/10.18653/v1/2020.emnlp-main.50} {Connecting
  the dots: Event graph schema induction with path language modeling}.
\newblock In \emph{Proceedings of the 2020 Conference on Empirical Methods in
  Natural Language Processing (EMNLP)}, pages 684--695, Online. Association for
  Computational Linguistics.

\bibitem[{Li et~al.(2018)Li, Ding, and Liu}]{Li2018ConstructingNarrativeGraph}
Zhongyang Li, Xiao Ding, and Ting Liu. 2018.
\newblock Constructing narrative event evolutionary graph for script event
  prediction.
\newblock In \emph{IJCAI}.

\bibitem[{Lin et~al.(2021{\natexlab{a}})Lin, Ma, Lin, Yang, Pradeep, and
  Nogueira}]{Lin_etal_SIGIR2021_Pyserini}
Jimmy Lin, Xueguang Ma, Sheng-Chieh Lin, Jheng-Hong Yang, Ronak Pradeep, and
  Rodrigo Nogueira. 2021{\natexlab{a}}.
\newblock {Pyserini}: A {Python} toolkit for reproducible information retrieval
  research with sparse and dense representations.
\newblock In \emph{Proceedings of the 44th Annual International ACM SIGIR
  Conference on Research and Development in Information Retrieval (SIGIR
  2021)}, pages 2356--2362.

\bibitem[{Lin et~al.(2021{\natexlab{b}})Lin, Yang, and
  Lin}]{lin-etal-2021-batch}
Sheng-Chieh Lin, Jheng-Hong Yang, and Jimmy Lin. 2021{\natexlab{b}}.
\newblock \href {https://doi.org/10.18653/v1/2021.repl4nlp-1.17} {In-batch
  negatives for knowledge distillation with tightly-coupled teachers for dense
  retrieval}.
\newblock In \emph{Proceedings of the 6th Workshop on Representation Learning
  for NLP (RepL4NLP-2021)}, pages 163--173, Online. Association for
  Computational Linguistics.

\bibitem[{Lin et~al.(2021{\natexlab{c}})Lin, Wang, Ji, Natarajan, and
  Liu}]{Lin2021}
Ying Lin, Han Wang, Heng Ji, Premkumar Natarajan, and Yang Liu.
  2021{\natexlab{c}}.
\newblock Personalized entity resolution with dynamic heterogeneous knowledge
  graph representations.
\newblock In \emph{Proc. ACL-IJCNLP2021 Workshop on e-Commerce and NLP}.

\bibitem[{Pichotta and Mooney(2014)}]{pichotta-mooney-2014-statistical}
Karl Pichotta and Raymond Mooney. 2014.
\newblock \href {https://doi.org/10.3115/v1/E14-1024} {Statistical script
  learning with multi-argument events}.
\newblock In \emph{Proceedings of the 14th Conference of the {E}uropean Chapter
  of the Association for Computational Linguistics}, pages 220--229,
  Gothenburg, Sweden. Association for Computational Linguistics.

\bibitem[{Pichotta and Mooney(2016)}]{pichotta-mooney-2016-statistical}
Karl Pichotta and Raymond Mooney. 2016.
\newblock \href {https://doi.org/10.18653/v1/W16-6003} {Statistical script
  learning with recurrent neural networks}.
\newblock In \emph{Proceedings of the Workshop on Uphill Battles in Language
  Processing: Scaling Early Achievements to Robust Methods}, pages 11--16,
  Austin, TX. Association for Computational Linguistics.

\bibitem[{Rae et~al.(2021)Rae, Borgeaud, Cai, Millican, Hoffmann, Song,
  Aslanides, Henderson, Ring, Young, Rutherford, Hennigan, Menick, Cassirer,
  Powell, van~den Driessche, Hendricks, Rauh, Huang, Glaese, Welbl, Dathathri,
  Huang, Uesato, Mellor, Higgins, Creswell, McAleese, Wu, Elsen, Jayakumar,
  Buchatskaya, Budden, Sutherland, Simonyan, Paganini, Sifre, Martens, Li,
  Kuncoro, Nematzadeh, Gribovskaya, Donato, Lazaridou, Mensch, Lespiau,
  Tsimpoukelli, Grigorev, Fritz, Sottiaux, Pajarskas, Pohlen, Gong, Toyama,
  de~Masson~d'Autume, Li, Terzi, Mikulik, Babuschkin, Clark, de~Las~Casas, Guy,
  Jones, Bradbury, Johnson, Hechtman, Weidinger, Gabriel, Isaac, Lockhart,
  Osindero, Rimell, Dyer, Vinyals, Ayoub, Stanway, Bennett, Hassabis,
  Kavukcuoglu, and Irving}]{Rae2021Gopher}
Jack~W. Rae, Sebastian Borgeaud, Trevor Cai, Katie Millican, Jordan Hoffmann,
  Francis Song, John Aslanides, Sarah Henderson, Roman Ring, Susannah Young,
  Eliza Rutherford, Tom Hennigan, Jacob Menick, Albin Cassirer, Richard Powell,
  George van~den Driessche, Lisa~Anne Hendricks, Maribeth Rauh, Po-Sen Huang,
  Amelia Glaese, Johannes Welbl, Sumanth Dathathri, Saffron Huang, Jonathan
  Uesato, John F.~J. Mellor, Irina Higgins, Antonia Creswell, Nathan McAleese,
  Amy Wu, Erich Elsen, Siddhant~M. Jayakumar, Elena Buchatskaya, David Budden,
  Esme Sutherland, Karen Simonyan, Michela Paganini, L.~Sifre, Lena Martens,
  Xiang~Lorraine Li, Adhiguna Kuncoro, Aida Nematzadeh, Elena Gribovskaya,
  Domenic Donato, Angeliki Lazaridou, Arthur Mensch, Jean-Baptiste Lespiau,
  Maria Tsimpoukelli, N.~K. Grigorev, Doug Fritz, Thibault Sottiaux, Mantas
  Pajarskas, Tobias Pohlen, Zhitao Gong, Daniel Toyama, Cyprien
  de~Masson~d'Autume, Yujia Li, Tayfun Terzi, Vladimir Mikulik, Igor
  Babuschkin, Aidan Clark, Diego de~Las~Casas, Aurelia Guy, Chris Jones, James
  Bradbury, Matthew~G. Johnson, Blake~A. Hechtman, Laura Weidinger, Iason
  Gabriel, William~S. Isaac, Edward Lockhart, Simon Osindero, Laura Rimell,
  Chris Dyer, Oriol Vinyals, Kareem~W. Ayoub, Jeff Stanway, L.~L. Bennett,
  Demis Hassabis, Koray Kavukcuoglu, and Geoffrey Irving. 2021.
\newblock Scaling language models: Methods, analysis \& insights from training
  gopher.
\newblock \emph{ArXiv}, abs/2112.11446.

\bibitem[{Reimers and Gurevych(2019)}]{reimers-gurevych-2019-sentence}
Nils Reimers and Iryna Gurevych. 2019.
\newblock \href {https://doi.org/10.18653/v1/D19-1410} {Sentence-{BERT}:
  Sentence embeddings using {S}iamese {BERT}-networks}.
\newblock In \emph{Proceedings of the 2019 Conference on Empirical Methods in
  Natural Language Processing and the 9th International Joint Conference on
  Natural Language Processing (EMNLP-IJCNLP)}, pages 3982--3992, Hong Kong,
  China. Association for Computational Linguistics.

\bibitem[{Rudinger et~al.(2015{\natexlab{a}})Rudinger, Rastogi, Ferraro, and
  Van~Durme}]{rudinger-etal-2015-script}
Rachel Rudinger, Pushpendre Rastogi, Francis Ferraro, and Benjamin Van~Durme.
  2015{\natexlab{a}}.
\newblock \href {https://doi.org/10.18653/v1/D15-1195} {Script induction as
  language modeling}.
\newblock In \emph{Proceedings of the 2015 Conference on Empirical Methods in
  Natural Language Processing}, pages 1681--1686, Lisbon, Portugal. Association
  for Computational Linguistics.

\bibitem[{Rudinger et~al.(2015{\natexlab{b}})Rudinger, Rastogi, Ferraro, and
  Van~Durme}]{rudinger2015script}
Rachel Rudinger, Pushpendre Rastogi, Francis Ferraro, and Benjamin Van~Durme.
  2015{\natexlab{b}}.
\newblock Script induction as language modeling.
\newblock In \emph{Proceedings of the 2015 Conference on Empirical Methods in
  Natural Language Processing}, pages 1681--1686.

\bibitem[{Sakaguchi et~al.(2021)Sakaguchi, Bhagavatula, Le~Bras, Tandon, Clark,
  and Choi}]{sakaguchi-etal-2021-proscript-partially}
Keisuke Sakaguchi, Chandra Bhagavatula, Ronan Le~Bras, Niket Tandon, Peter
  Clark, and Yejin Choi. 2021.
\newblock \href {https://doi.org/10.18653/v1/2021.findings-emnlp.184}
  {pro{S}cript: Partially ordered scripts generation}.
\newblock In \emph{Findings of the Association for Computational Linguistics:
  EMNLP 2021}, pages 2138--2149, Punta Cana, Dominican Republic. Association
  for Computational Linguistics.

\bibitem[{Sancheti and Rudinger(2022)}]{sancheti-rudinger-2022-large}
Abhilasha Sancheti and Rachel Rudinger. 2022.
\newblock \href {https://doi.org/10.18653/v1/2022.starsem-1.1} {What do large
  language models learn about scripts?}
\newblock In \emph{Proceedings of the 11th Joint Conference on Lexical and
  Computational Semantics}, pages 1--11, Seattle, Washington. Association for
  Computational Linguistics.

\bibitem[{Schank and Abelson(1975)}]{schank1975scripts}
Roger~C Schank and Robert~P Abelson. 1975.
\newblock Scripts, plans, and knowledge.
\newblock In \emph{IJCAI}, volume~75, pages 151--157.

\bibitem[{Wang et~al.(2022)Wang, Wei, Schuurmans, Le, Chi, and
  Zhou}]{Wang2022RationaleAugmentedEI}
Xuezhi Wang, Jason Wei, Dale Schuurmans, Quoc Le, Ed~Chi, and Denny Zhou. 2022.
\newblock Rationale-augmented ensembles in language models.
\newblock \emph{ArXiv}, abs/2207.00747.

\bibitem[{Wang et~al.(2017)Wang, Zhang, and Chang}]{wang-etal-2017-integrating}
Zhongqing Wang, Yue Zhang, and Ching-Yun Chang. 2017.
\newblock \href {https://doi.org/10.18653/v1/D17-1006} {Integrating order
  information and event relation for script event prediction}.
\newblock In \emph{Proceedings of the 2017 Conference on Empirical Methods in
  Natural Language Processing}, pages 57--67, Copenhagen, Denmark. Association
  for Computational Linguistics.

\bibitem[{Weber et~al.(2018)Weber, Balasubramanian, and
  Chambers}]{Weber2018EventRW}
Noah Weber, Niranjan Balasubramanian, and Nathanael Chambers. 2018.
\newblock Event representations with tensor-based compositions.
\newblock In \emph{AAAI}.

\bibitem[{Wei et~al.(2022)Wei, Bosma, Zhao, Guu, Yu, Lester, Du, Dai, and
  Le}]{Wei2021FLAN}
Jason Wei, Maarten Bosma, Vincent Zhao, Kelvin Guu, Adams~Wei Yu, Brian Lester,
  Nan Du, Andrew~M. Dai, and Quoc~V. Le. 2022.
\newblock Finetuned language models are zero-shot learners.
\newblock \emph{ICLR}.

\bibitem[{Winkler(1990)}]{winkler1990string}
William~E Winkler. 1990.
\newblock String comparator metrics and enhanced decision rules in the
  fellegi-sunter model of record linkage.

\bibitem[{Zhang et~al.(2020)Zhang, Lyu, and
  Callison-Burch}]{zhang-etal-2020-reasoning}
Li~Zhang, Qing Lyu, and Chris Callison-Burch. 2020.
\newblock \href {https://doi.org/10.18653/v1/2020.emnlp-main.374} {Reasoning
  about goals, steps, and temporal ordering with {W}iki{H}ow}.
\newblock In \emph{Proceedings of the 2020 Conference on Empirical Methods in
  Natural Language Processing (EMNLP)}, pages 4630--4639, Online. Association
  for Computational Linguistics.

\bibitem[{Zhou et~al.(2021)Zhou, Richardson, Ning, Khot, Sabharwal, and
  Roth}]{zhou-etal-2021-temporal}
Ben Zhou, Kyle Richardson, Qiang Ning, Tushar Khot, Ashish Sabharwal, and Dan
  Roth. 2021.
\newblock \href {https://doi.org/10.18653/v1/2021.naacl-main.107} {Temporal
  reasoning on implicit events from distant supervision}.
\newblock In \emph{Proceedings of the 2021 Conference of the North American
  Chapter of the Association for Computational Linguistics: Human Language
  Technologies}, pages 1361--1371, Online. Association for Computational
  Linguistics.

\bibitem[{Zhou et~al.(2022{\natexlab{a}})Zhou, He, Ma, Berg-Kirkpatrick, and
  Neubig}]{Zhou2022PromptConsistency}
Chunting Zhou, Junxian He, Xuezhe Ma, Taylor Berg-Kirkpatrick, and Graham
  Neubig. 2022{\natexlab{a}}.
\newblock Prompt consistency for zero-shot task generalization.
\newblock \emph{ArXiv}, abs/2205.00049.

\bibitem[{Zhou et~al.(2022{\natexlab{b}})Zhou, Zhang, Yang, Lyu, Yin,
  Callison-Burch, and Neubig}]{zhou-etal-2022-show}
Shuyan Zhou, Li~Zhang, Yue Yang, Qing Lyu, Pengcheng Yin, Chris Callison-Burch,
  and Graham Neubig. 2022{\natexlab{b}}.
\newblock \href {https://doi.org/10.18653/v1/2022.acl-long.214} {Show me more
  details: Discovering hierarchies of procedures from semi-structured web
  data}.
\newblock In \emph{Proceedings of the 60th Annual Meeting of the Association
  for Computational Linguistics (Volume 1: Long Papers)}, pages 2998--3012,
  Dublin, Ireland. Association for Computational Linguistics.

\end{thebibliography}
